\pdfoutput=1

\documentclass[11pt]{article}

\usepackage[utf8]{inputenc} % Input encoding
\usepackage[T2A,T1]{fontenc} % Font encoding: T2A for Russian, T1 for Western
\usepackage[russian,english]{babel} % Language support

\usepackage[final]{acl}
\usepackage{times}
\usepackage{inconsolata} % Typewriter font
\usepackage{microtype}   % Improved typography

\usepackage[table,dvipsnames]{xcolor} 
\usepackage{graphicx}
\usepackage{pgfplots}
\usepgfplotslibrary{groupplots}
\pgfplotsset{compat=1.18}

\usepackage{amsmath}
\usepackage{amssymb}
\usepackage{latexsym}
\usepackage{pifont} % For checkmarks/xmarks

\usepackage{booktabs}   % Professional table rules
\usepackage{multirow}   % Spanning rows
\usepackage{multicol}   % Spanning columns
\usepackage{longtable}  % Tables spanning pages
\usepackage{adjustbox}  % Resizing tables
\usepackage{caption}
\usepackage{float}

\usepackage{url}
\usepackage[many]{tcolorbox}
\tcbuselibrary{listings}

\definecolor{darkgreen}{rgb}{0.0,0.5,0.0}

\title{\textsc{IchthyoNoma}: Nomenclature and Context Sensitivity of Zero-Shot Biological Vision--Language Models for Bangladeshi Freshwater Fish Recognition}
\author{
 \textbf{
Nazim-E-Alam \textsuperscript{1}\thanks{Equal contribution.}}, 
 \textbf{Tarek Rahman\textsuperscript{2,3,*}\textsuperscript{$\ddagger$} \thanks{Correspondence: \href{mailto:trahman221182@bscse.uiu.ac.bd}{trahman221182@bscse.uiu.ac.bd}}}, 
 \textbf{Md Kishor Morol\textsuperscript{3}\thanks{Equal supervision.}}
\\
\textsuperscript{1}American International University Bangladesh,
 \textsuperscript{2}United International University,
 \\
 \textsuperscript{3} ELITE Research Lab, Queens, New York, USA \\
\small{
}
}

\begin{document}
\maketitle
\begin{abstract}
Zero-shot vision--language models (VLMs) are increasingly used as training-free species recognizers, but reported accuracy can reflect more than visual species knowledge. We audit CLIP, BioCLIP, BioCLIP2, and a multilingual Jina CLIP v2 control on seven freshwater-fish categories from two Bangladeshi sources (10,321 images). BioCLIP2 reaches 72.36\% on BFF-15 with English common names and 68.91\% on SylFishBD with scientific names, versus 25.15\% and 14.40\% for generic CLIP. BioCLIP2 Bengali prompts are near chance in balanced accuracy (14.22--14.29\%); Jina partially recovers Bengali discrimination to 21.89\% and 16.36\%, but bare Bengali names return to 14.29\% on both sources. Paired SylFishBD interventions show no significant weak-blur effect, modest losses from stronger blur/gray masking, a larger white-mask artifact, and strong species dependence. Zero-shot biological VLM scores therefore jointly reflect biological specialization, multilingual alignment, nomenclature, prompt formulation, and context.
All the codes and data are available at GitHub:  \url{https://github.com/NazimRiyadh/IchthyoNoma}.
\end{abstract}
\begin{figure*}[t]
\centering
\includegraphics[width=0.97\textwidth]{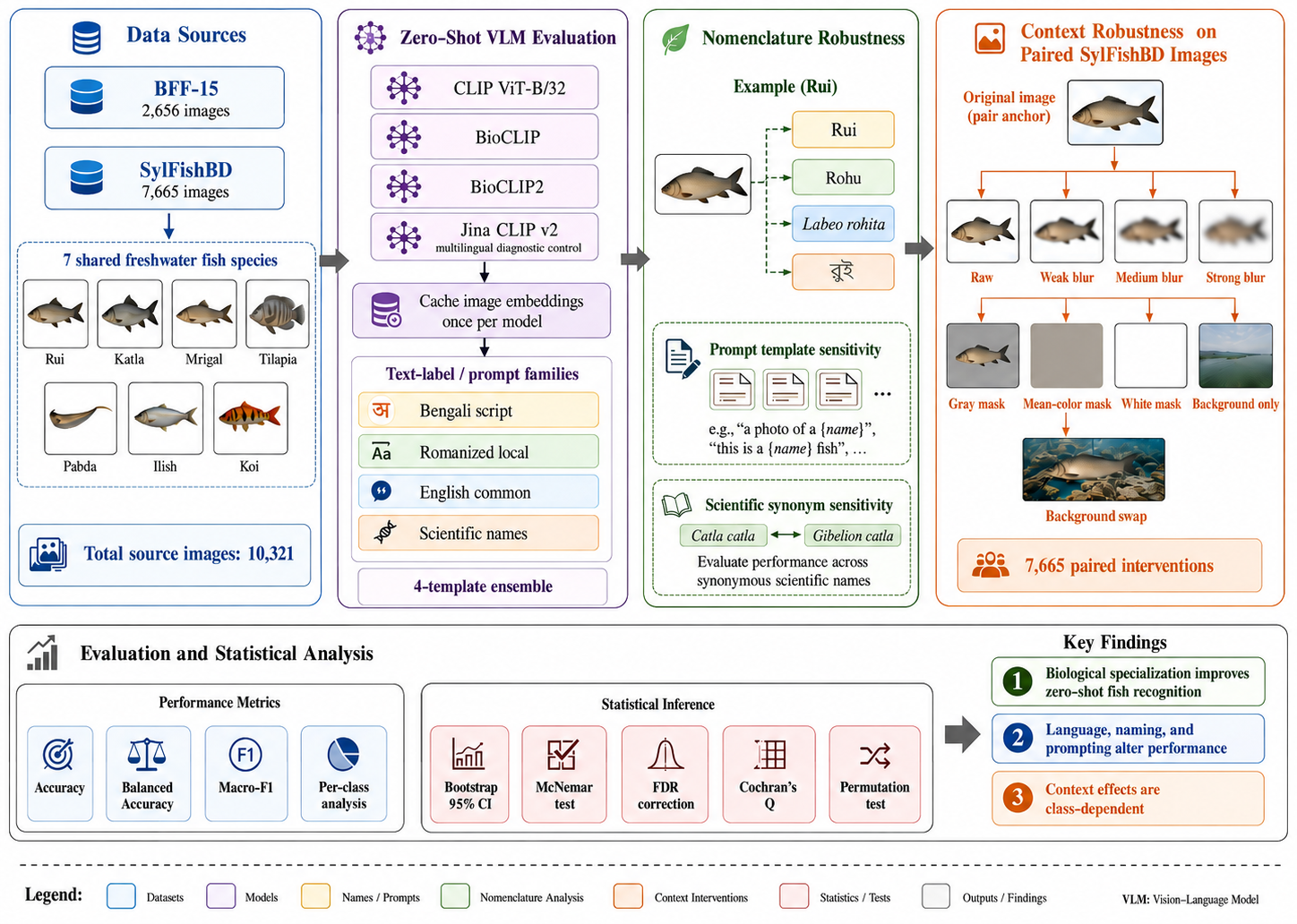}
\caption{Overview of the study framework. Seven shared freshwater fish species from BFF-15 and SylFishBD are evaluated using frozen CLIP, BioCLIP, and BioCLIP2 models, with Jina CLIP v2 included as a multilingual control. The framework assesses sensitivity to label language, nomenclature, prompt formulation, and scientific synonyms, together with paired visual-context interventions on SylFishBD. Performance is quantified using standard classification metrics, per-class analysis, and statistical significance testing to characterize the effects of biological specialization, textual formulation, and visual context on zero-shot fish recognition.}
\label{fig:overview}
\end{figure*}
\section{Introduction}\label{intro}
Automated species recognition supports biodiversity monitoring, fisheries, aquaculture, market inspection, and increasingly automated grading and robotic sorting systems \citep{rahmanfishnetbot}. Fish recognition is difficult because related taxa may differ subtly while pose, illumination, occlusion, and background vary. Conventional systems therefore rely on labeled data and supervised transfer learning \citep{shah2019fishpak,xu2021seresnet,das2024bdfreshwater}, but high in-domain accuracy does not establish transferable biological knowledge.
Contrastive vision--language pre-training offers a training-free alternative. CLIP enables zero-shot image--text classification \citep{radford2021clip}, while BioCLIP and BioCLIP2 add biologically structured pre-training for fine-grained recognition \citep{stevens2024bioclip,gu2025bioclip2}. Fish work has also explored language supervision and training-free multimodal recognition \citep{dai2024clipfssc,jose2026zeroshot}.
For VLMs, the class label is part of the classifier: small wording changes can alter CLIP predictions \citep{zhou2022coop,zhou2022cocoop}. Biology amplifies this issue because one organism may have vernacular, English-common, scientific, historical, or synonymous names; scientific binomials are not always optimal \citep{parashar2023scientific}, and multilingual VLMs show large cross-language gaps \citep{geigle2024babel,chung2026uclip}. In Bangladesh, success through English or Latin strings but failure through Bengali names can overstate local accessibility.
Visual context introduces a parallel ambiguity: classifiers can exploit backgrounds and shortcuts \citep{geirhos2020shortcut,xiao2021backgrounds}, while CLIP can retain domain-sensitive information \citep{wen2025diverse}.

We therefore ask: \emph{how stable is zero-shot biological recognition when the model, language, nomenclature, prompt, or context changes?}
Our contributions are:
\begin{itemize}
  \item A cross-model, cross-source zero-shot benchmark of CLIP, BioCLIP, and BioCLIP2 on seven Bangladeshi freshwater-fish categories;
  \item A nomenclature audit spanning Romanized vernacular, Bengali script, English-common and scientific names, prompt templates, and scientific synonyms, augmented with a multilingual Jina CLIP v2 control that separates multilingual alignment from biological specialization;
  \item Paired context interventions on 7,665 SylFishBD images using graded blur, neutral masks, cropping, background-only views, and cross-species background swaps, with paired inference and multiple-testing correction.
\end{itemize}

\section{Related Work}\label{back}
Prior work relevant to this study spans prompt and nomenclature sensitivity, biological and multilingual vision-language models, and fish recognition under varying visual context. We review these directions to position our evaluation of zero-shot recognition across language, naming, model specialization, and contextual variation.

\paragraph{Prompt and nomenclature robustness.}
CLIP performs zero-shot classification by constructing class prototypes from natural-language
descriptions rather than learning task-specific classification weights
\citep{radford2021clip}. Consequently, the textual specification of a category becomes part of
the classifier itself. Prior work on CoOp and CoCoOp demonstrates that modifying prompt context
can substantially alter downstream recognition behavior
\citep{zhou2022coop,zhou2022cocoop}. This sensitivity is particularly relevant for fine-grained
biological recognition, where the same organism may be represented by vernacular names,
common English names, scientific binomials, or historical taxonomic synonyms.
Parashar et al. showed that common English names can substantially outperform scientific names
for zero-shot species recognition with generic CLIP, indicating that taxonomically precise labels
are not necessarily the most effective textual interface
\citep{parashar2023scientific}. These findings motivate our treatment of label language,
nomenclature, exact taxonomic strings, and prompt templates as explicit evaluation variables
rather than fixed preprocessing choices.

\paragraph{Biological and ecological vision--language models.}
Generic VLMs are trained primarily on broad web-scale image--text data and may therefore lack
the fine-grained biological structure required for species-level recognition. BioCLIP addresses
this limitation through biologically structured contrastive pre-training, while BioCLIP2 expands
this approach with substantially greater biological and taxonomic coverage
\citep{stevens2024bioclip,gu2025bioclip2}. Related work has also explored multimodal learning
specifically for aquatic species. CLIP-FSSC uses natural-language supervision for transferable
fish and shrimp species classification and demonstrates the potential of VLM-based representations
for recognition without conventional downstream annotation
\citep{dai2024clipfssc}. More broadly, TaxaBind constructs a unified ecological representation
across species imagery, taxonomic text, geographic location, satellite imagery, audio, and
environmental information, demonstrating zero-shot capabilities for ecological tasks including
species classification \citep{sastry2025taxabind}. These studies establish the value of biologically
informed multimodal representations; our work instead focuses on how stable a frozen biological
VLM remains when the textual interface and visual context are systematically altered.

\paragraph{Multilingual vision--language alignment.}
Biological specialization and multilingual capability represent distinct dimensions of VLM
performance. Most zero-shot vision--language benchmarks have historically emphasized English,
which can obscure substantial degradation when equivalent concepts are expressed in other
languages. Babel-ImageNet evaluates multilingual vision--language representations across
100 languages and reports substantial cross-language performance variation, with particularly
large gaps for lower-resource languages \citep{geigle2024babel}. Parameter-efficient approaches
such as uCLIP further demonstrate that multilingual alignment can be improved without rebuilding
the entire vision--language model \citep{chung2026uclip}. In the present study, Bengali is not
treated merely as another prompt variant: it provides a test of whether biological visual
specialization is accompanied by an accessible native-language textual interface. We therefore
include multilingual Jina CLIP v2 \citep{koukounas2024jinaclipv2} as a diagnostic control,
allowing multilingual text alignment to be examined separately from fine-grained biological
specialization.

\paragraph{Regional fish recognition and visual context.}
Bangladesh-specific fish datasets provide an important setting for evaluating whether biological
VLMs transfer to locally relevant taxa. SylFishBD provides freshwater-fish imagery together with
SAM-derived instance masks \citep{absar2026sylfish,kirillov2023sam}, while BFF-15 provides a
separate collection of Bangladeshi freshwater fish images \citep{bff15}. Recent work has achieved
strong in-domain fish recognition through supervised or self-supervised adaptation
\citep{siam2025freshwater}; however, such results do not directly establish training-free
transferability because the model has been adapted to the target recognition task. In contrast,
our evaluation keeps all model weights and prompts frozen.

Visual context introduces an additional source of uncertainty. Recognition systems can exploit
background correlations as shortcuts rather than relying exclusively on the target organism,
and prior work shows that background information alone can retain substantial class-discriminative
signal \citep{xiao2021backgrounds}. More recent studies of background bias similarly show that
spurious correlations between foreground categories and contextual regions can reduce
generalization robustness \citep{bassi2024background}. These observations motivate our paired
context interventions, including graded blur, neutral-color masking, cropping, background-only
views, and cross-species background swaps. Rather than interpreting any single synthetic
transformation as causal evidence, we use a ladder of interventions to test whether zero-shot
fish predictions remain stable as contextual information is progressively modified.

\begin{table*}[t]
\caption{Shared seven-species benchmark across BFF-15 and SylFishBD, including class nomenclature and image counts. Each SylFishBD image is paired with a segmentation mask.}
\label{tab:data}
\centering
\small
\begin{tabular}{l l l l rr}
\toprule
Class & Romanized & English & Scientific & BFF & Syl \\
\midrule
Rui & Rui & Rohu & \emph{L. rohita} & 514 & 1670 \\
Katla & Katla & Catla & \emph{C. catla} & 427 & 1133 \\
Mrigal & Mrigal & Mrigal carp & \emph{C. cirrhosus} & 317 & 1293 \\
Tilapia & Tilapia & Nile tilapia & \emph{O. niloticus} & 383 & 1326 \\
Pabda & Pabda & Pabda catfish & \emph{O. pabda} & 348 & 862 \\
Ilish & Ilish & Hilsa & \emph{T. ilisha} & 233 & 789 \\
Koi & Koi & Climbing perch & \emph{A. testudineus} & 434 & 592 \\
\midrule
Total &&&& 2656 & 7665 \\
\bottomrule
\end{tabular}
\end{table*}
\section{Methodology}
This section outlines the experimental setup for evaluating zero-shot fish recognition across datasets, models, naming schemes, languages, and visual contexts. It covers dataset harmonization, frozen zero-shot classification, the Jina CLIP v2 multilingual control, context interventions, and statistical evaluation using accuracy, balanced accuracy, macro-F1, bootstrap confidence intervals, and paired significance tests.
\subsection{Data harmonization}
We use seven categories shared by BFF-15 and SylFishBD: Rui, Katla, Mrigal, Tilapia, Pabda, Ilish, and Koi. BFF-15 contributes 2,656 images and SylFishBD 7,665 (10,321 total); every selected SylFishBD image has a binary mask. Table~\ref{tab:data} gives counts/nomenclature. All models remain frozen, so no train/validation/test split is required.

SHA-256 found no exact cross-source duplicates; a 64-bit perceptual-hash search flagged 73 low-distance pairs ($\leq 6$ bits) for review. Because similar fish can yield similar hashes, we flag rather than automatically exclude them and describe the collections as two public sources, not fully independent sources.

\subsection{Frozen zero-shot classification and prompt construction}
We evaluate CLIP ViT-B/32 (OpenAI), BioCLIP, and BioCLIP2. For normalized image embedding $f_I(x)$ and class name $n_c$, we instantiate four templates---``a photo/image/photograph/specimen of $n_c$, a fish species''---normalize each text embedding, and form the normalized mean prototype
\begin{equation}
 t_c = \frac{\frac{1}{K}\sum_{k=1}^K \hat f_T(p_k(n_c))}{\left\|\frac{1}{K}\sum_{k=1}^K \hat f_T(p_k(n_c))\right\|_2}, \qquad K=4.
\end{equation}
Prediction is $\hat y=\arg\max_c f_I(x)^\top t_c$. Cached image embeddings ensure nomenclature experiments change only the text-side classifier.

We test Romanized, English-common, and scientific names, plus Bengali script for BioCLIP2. Templates are also evaluated individually. Scientific-name controls substitute \emph{Catla catla} with \emph{Gibelion catla} or \emph{Labeo catla}, and \emph{Cirrhinus cirrhosus} with \emph{C. mrigala}; these are nomenclatural variants, not claims of uniquely correct taxonomy.

\subsection{Multilingual Bengali control}
To test whether BioCLIP2's Bengali failure reflects text alignment, we add frozen Jina CLIP v2 \citep{koukounas2024jinaclipv2}. The same 10,321 images/classes use Bengali, Romanized, English-common, and scientific labels with the same four-template ensemble and a 512-dimensional Matryoshka-truncated representation; no benchmark tuning is performed.

A secondary \emph{name-only} control embeds only Bengali class names, testing whether multilingual support alone suffices or interacts with prompt framing. This directly tests whether BioCLIP2's near-chance Bengali result is partly a text-alignment limitation rather than a property of Bengali nomenclature itself. Jina is not used for context interventions; it is diagnostic for multilingual alignment versus biological specialization.

\subsection{Paired context interventions}
SylFishBD masks preserve foreground pixels while altering background. We apply Gaussian blur ($\sigma=0.01,0.03,0.06$ times the shorter dimension), white/gray/mean-color masks, a 10\%-expanded tight crop, foreground removal with Telea inpainting, and cross-species background swaps. These are stress tests, not perfect causal isolation, because synthetic transformations introduce distribution shift.

\subsection{Metrics and inference}
We report accuracy, balanced accuracy, macro-F1, and per-class effects. Class-stratified bootstrap CIs use 1,000 resamples for the primary benchmark and 2,000 for robustness/multilingual controls. Paired comparisons use exact two-sided McNemar tests and bootstrap accuracy differences with Benjamini--Hochberg FDR correction. Cochran's $Q$ tests context conditions; donor-follow uses 5,000 label permutations. Seed is 42.

\begin{table*}[t]
\caption{Primary zero-shot classification performance across models, prompt types, and datasets. Values are reported as accuracy / macro-F1 (\%); bold indicates the highest accuracy for each dataset.}
\label{tab:primary}
\centering
\small
\begin{tabular}{llcc}
\toprule
Model & Prompt & BFF-15 & SylFishBD \\
\midrule
CLIP ViT-B/32 & Romanized & 10.96 / 5.64 & 16.95 / 13.15 \\
 & English & 25.15 / 20.03 & 23.80 / 22.05 \\
 & Scientific & 15.40 / 6.97 & 14.40 / 7.21 \\
\midrule
BioCLIP & Romanized & 25.26 / 18.28 & 27.93 / 16.63 \\
 & English & 54.89 / 51.61 & 52.26 / 46.64 \\
 & Scientific & 47.82 / 43.34 & 58.88 / 57.45 \\
\midrule
BioCLIP2 & Romanized & 35.77 / 25.98 & 37.56 / 25.19 \\
 & English & \textbf{72.36} / 67.33 & 64.59 / 64.30 \\
 & Scientific & 69.99 / 68.85 & \textbf{68.91} / 69.68 \\
\bottomrule
\end{tabular}
\end{table*}

\section{Results and Discussion}
This section presents the main experimental findings across model specialization, nomenclature and prompt sensitivity, multilingual alignment, and visual-context robustness. First, the zero-shot performance of generic CLIP is compared with the biology-specialized BioCLIP and BioCLIP2 models on BFF-15 and SylFishBD, showing a substantial advantage for biologically specialized pre-training. Next, the analysis examines how Romanized, English-common, scientific, and Bengali class names, as well as different prompt templates and scientific-name variants, affect recognition performance. Jina CLIP v2 is subsequently used as a multilingual control to assess whether BioCLIP2’s poor performance with Bengali prompts stems from limitations in cross-language text alignment. While Jina improves Bengali discrimination to some extent, its overall fine-grained fish recognition performance remains substantially below that of BioCLIP2.
\begin{figure*}[ht]
\centering
\includegraphics[width=1\textwidth]{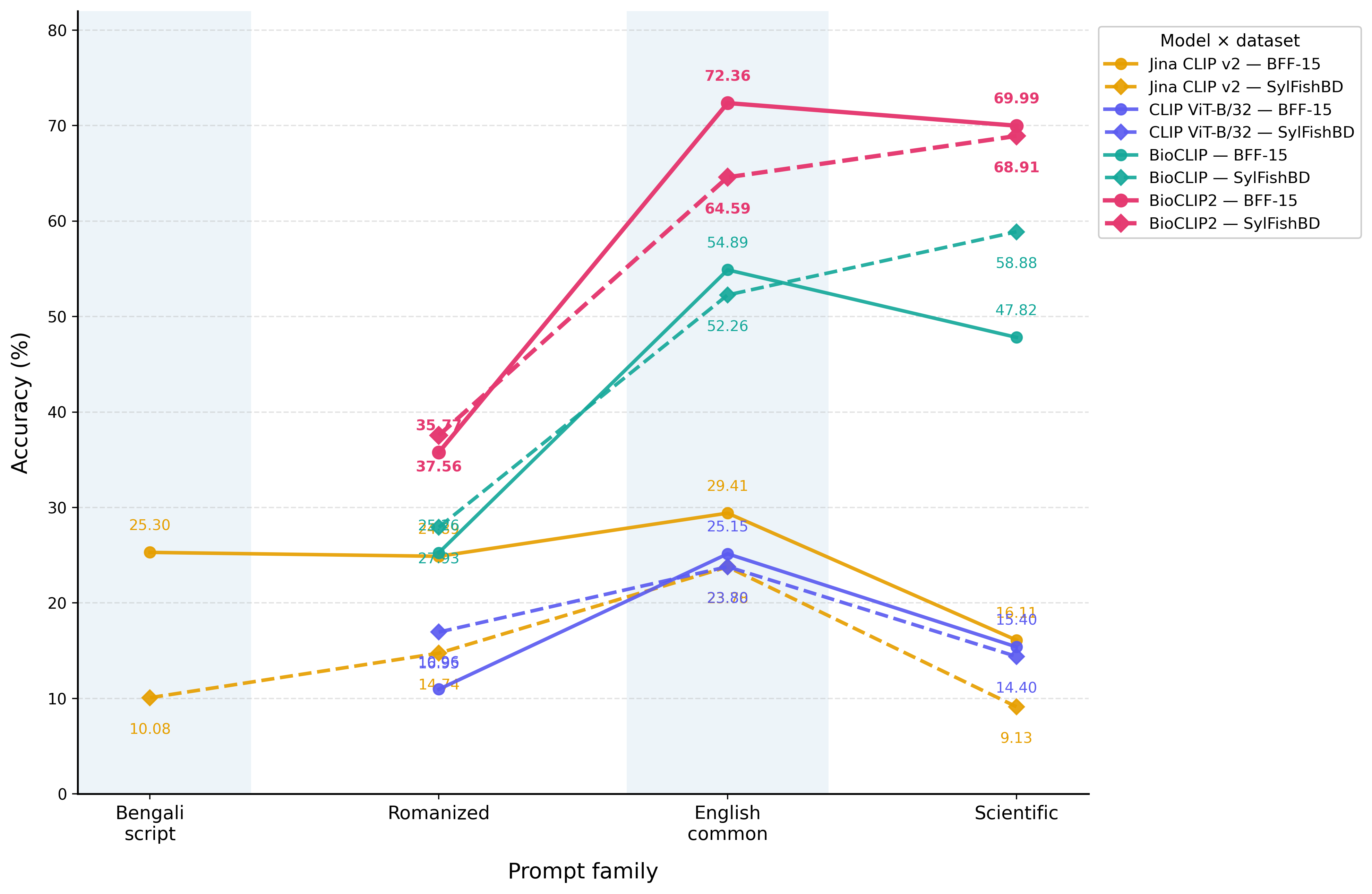}
\caption{Cross-model zero-shot accuracy across prompt families and datasets. CLIP ViT-B/32, BioCLIP, and BioCLIP2 are compared under Romanized, English-common, and scientific labels, while Jina CLIP v2 additionally includes Bengali-script prompting. Solid and dashed lines denote BFF-15 and SylFishBD, respectively.}
\label{fig:prompt_profile}
\end{figure*}

Finally, paired context interventions on SylFishBD examine the effects of blur, masking, cropping, background removal, and background swapping. These experiments demonstrate that context sensitivity is intervention- and species-dependent rather than a uniform background effect.

\subsection{Biology-specialized VLMs outperform the tested generic CLIP baseline}
Table~\ref{tab:primary} shows a large gap between generic and biology-specialized VLMs. With scientific names, CLIP scores 15.40\%/14.40\% on BFF-15/SylFishBD, BioCLIP 47.82\%/58.88\%, and BioCLIP2 69.99\%/68.91\%. With English names BioCLIP2 reaches 72.36\%/64.59\%. Relative to BioCLIP, BioCLIP2 gains 22.18 and 10.03 points under scientific prompting (paired 95\% CIs [19.73, 24.66] and [8.91, 11.17], both $q<0.001$). Because checkpoints also differ in scale/training recipe, we do not attribute the full gap solely to specialization.

\subsection{Nomenclature and prompt formulation are part of the classifier}
With identical BioCLIP2 image embeddings, BFF-15 scores 35.77\%/72.36\%/69.99\% for Romanized/English/scientific names; SylFishBD scores 37.56\%/64.59\%/68.91\%. Scientific terminology is therefore not universally optimal.

Bengali prompts yield 16.27\%/7.74\% accuracy and 14.22\%/14.29\% balanced accuracy, essentially seven-class chance (14.29\%). Thus these Bengali fish-name prompts provide little discrimination for the tested BioCLIP2 text encoder.

Across four individual scientific templates, BioCLIP2 spans 56.06--77.07\% on BFF-15 and 61.44--79.71\% on SylFishBD. Replacing \emph{Catla catla} by \emph{Gibelion catla} raises SylFishBD accuracy 5.14 points ($q<0.001$), while \emph{Labeo catla} lowers it 3.46 points and lowers BFF-15 by 8.58 points. Replacing \emph{Cirrhinus cirrhosus} by \emph{C. mrigala} improves both sources by about two points. Thus ``scientific prompt'' is not fully specified without the exact string/template.
\begin{table*}[ht]
\caption{Jina CLIP v2 multilingual control for evaluating Bengali and cross-language prompt sensitivity. Values are reported as accuracy / balanced accuracy / macro-F1 (\%).}
\label{tab:jina}
\centering
\small
\begin{tabular}{lcc}
\toprule
Prompt family & BFF-15 & SylFishBD \\
\midrule
Bengali script & 25.30 / 21.89 / 16.31 & 10.08 / 16.36 / 6.04 \\
Romanized & 24.89 / 23.93 / 19.05 & 14.74 / 14.78 / 12.06 \\
English common & 29.41 / 28.43 / 17.87 & 23.78 / 18.01 / 15.62 \\
Scientific & 16.11 / 15.96 / 13.56 & 9.13 / 10.27 / 9.75 \\
\bottomrule
\end{tabular}
\end{table*}
\subsection{Multilingual alignment partially recovers Bengali, but does not replace biological specialization}
Under the same four-template ensemble, Jina CLIP v2 reaches 21.89\% Bengali balanced accuracy on BFF-15 (95\% CI [20.77, 23.01]) and 16.36\% on SylFishBD ([15.64, 17.05]), versus BioCLIP2's 14.22\%/14.29\%. This descriptive control suggests that the near-chance BioCLIP2 result is partly model/interface-specific.  Table~\ref{tab:jina} summarizes Jina CLIP v2 performance across the four
label families on both datasets. And Figure~\ref{fig:prompt_profile} provides a unified comparison of zero-shot accuracy
across prompt families, models, and datasets, highlighting the contrasting effects
of biological specialization and multilingual alignment.

Recovery remains incomplete: English exceeds Bengali accuracy by 4.10 points on BFF-15 (CI [2.86, 5.35], $q<10^{-8}$) and 13.70 on SylFishBD ([12.77, 14.66], $q<10^{-98}$). Bengali and Romanized are indistinguishable on BFF-15 ($q=0.617$), while Romanized exceeds Bengali by 4.66 points on SylFishBD ($q<10^{-19}$).

Jina's English accuracy (29.41\%/23.78\%) remains far below BioCLIP2 (72.36\%/64.59\%), so multilingual alignment does not replace biological specialization. Bare Bengali names also collapse to exactly 14.29\% balanced accuracy on both sources, showing an interaction with prompt framing. Together, the controls expose a trade-off: stronger multilingual alignment can recover Bengali signal without matching BioCLIP2's fine-grained biological accuracy. We therefore interpret Jina as an interface diagnostic, not evidence about Bengali understanding in general.

\subsection{Context sensitivity is real but intervention-dependent}
The white-mask comparison drops BioCLIP2 from 68.91\% to 60.52\% ($-8.39$ points), but the intervention ladder qualifies this effect (Table~\ref{tab:context}). Weak blur is nonsignificant ($-0.47$; CI [$-1.34$, 0.46], $q=0.347$); medium/strong blur reduce accuracy 2.18/2.87 points and gray/mean-color masking about 3.8 points (all $q<0.001$). Thus white masking adds substantial distribution shift.
\begin{figure*}[!ht]
\centering
\includegraphics[width=.75\textwidth]{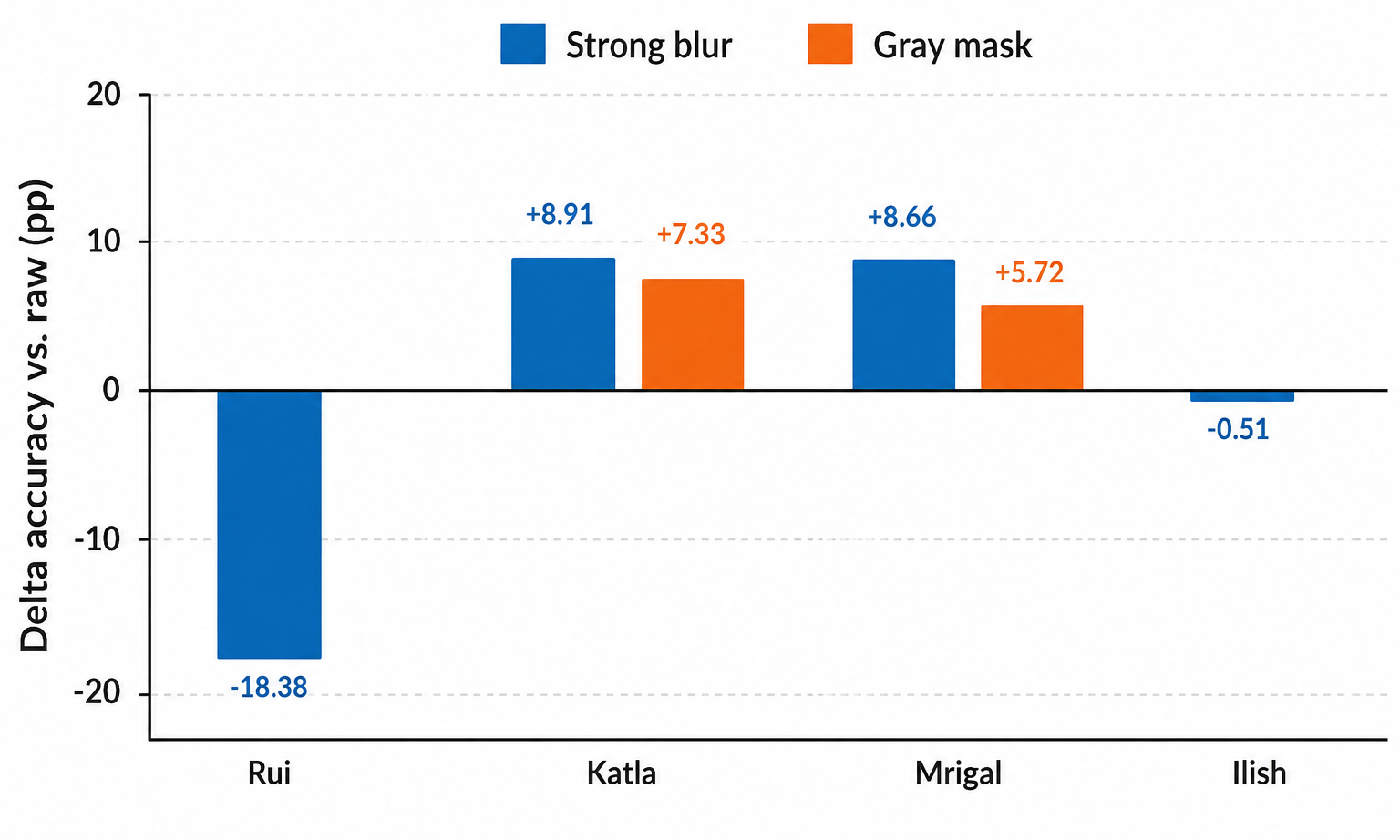}
\caption{Class-conditional changes from raw SylFishBD. Strong blur helps Katla/Mrigal but sharply hurts Rui; Ilish is nearly unchanged. Gray-mask values are shown where exact effects are available.}
\label{fig:classcontext}
\end{figure*}
\begin{table*}[!ht]
\caption{BioCLIP2 performance under paired context interventions on 7,665 SylFishBD images. $\Delta$ denotes the accuracy change from the raw condition, with paired-bootstrap 95\% confidence intervals shown in brackets.}
\label{tab:context}
\centering
\small
\begin{tabular}{lrrl}
\toprule
Condition & Acc. & $\Delta$ (pp) & FDR $q$ \\
\midrule
Raw & 68.91 & -- & -- \\
Weak blur & 68.44 & $-0.47$ [$-1.34$, 0.46] & 0.347 \\
Medium blur & 66.73 & $-2.18$ [$-3.11$, $-1.23$] & $<0.001$ \\
Strong blur & 66.04 & $-2.87$ [$-3.81$, $-1.94$] & $<0.001$ \\
Gray mask & 65.11 & $-3.80$ [$-4.79$, $-2.83$] & $<0.001$ \\
Mean-color mask & 65.10 & $-3.81$ [$-4.76$, $-2.81$] & $<0.001$ \\
White mask & 60.52 & $-8.39$ [$-9.56$, $-7.25$] & $<0.001$ \\
\bottomrule
\end{tabular}
\end{table*}

Aggregate means conceal class dependence (Fig.~\ref{fig:classcontext}). Strong blur improves Katla/Mrigal by 8.91/8.66 points, hurts Rui by 18.38, and barely changes Ilish ($-0.51$). Gray masking shows the same heterogeneity: Katla/Mrigal improve (+7.33/+5.72), while Rui, Koi, Tilapia, and Pabda decline.

Supporting controls reject simple background-label explanations. Background-only images retain 17.90\% accuracy (21.48\% balanced; $-51.01$ points, CI [$-52.17$, $-49.84$]), confirming dominant foreground information. Cross-species swaps reduce accuracy to 56.76\% ($-12.15$; CI [$-13.24$, $-11.06$]), yet donor-follow is only 8.88\% versus a 15.10\% permutation-null mean. Tight cropping reaches 55.15\%, and five mask-geometry properties are nonsignificant after FDR ($q\ge0.097$).

\subsection{Implications and limitations}
Three implications follow. Biology-specialized VLMs outperform the tested generic CLIP baseline, although scale/training recipe also differ. Zero-shot accuracy is not an image-encoder property alone: language, exact taxonomic string, and prompt frame instantiate the classifier, as the multilingual/name-only controls demonstrate. Context robustness likewise requires a ladder of interventions rather than one synthetic mask. Practically, reporting only the best English/scientific prompt can overstate accessibility when a strong visual model exposes a weak native-language interface.

Limitations include only seven categories/two sources, unknown pre-training exposure to related imagery, one multilingual control with a 512-D truncated representation, Bengali results specific to the tested names/templates, and synonym tests limited to Katla/Mrigal. One multilingual model cannot establish a general property of multilingual VLMs. Context interventions change image distribution as well as information content, so conclusions rely on agreement across blur, masks, crop, background-only, and swap controls rather than perfect causal decomposition.

For regional biological deployment, we recommend reporting zero-shot performance under at least English-common, scientific, and locally used names rather than selecting a single best prompt. Native-language interfaces should be evaluated separately, and context robustness should be tested with multiple interventions instead of a single masking condition.

\section{Conclusion}
We systematically evaluate zero-shot freshwater-fish recognition across generic, biology-specialized, and multilingual vision--language models. BioCLIP2 achieves the strongest overall performance, but its predictions remain sensitive to nomenclature, language, prompt formulation, taxonomic synonyms, and visual context. Jina CLIP v2 partially improves Bengali-language discrimination, yet remains substantially weaker for fine-grained biological recognition and falls to chance-level performance when evaluated with bare Bengali class names. Paired context interventions further reveal that mild visual degradation produces limited average effects, whereas stronger masking introduces larger distribution-shift artifacts and pronounced species-specific variation. Overall, the findings indicate that biological specialization and multilingual alignment address distinct sources of error. Zero-shot biological benchmarks should therefore explicitly evaluate language, exact nomenclature, prompt formulation, and contextual robustness rather than reporting performance under a single textual or visual configuration.
\bibliography{custom}

\end{document}